\documentclass[final,5p,times,twocolumn]{elsarticle}
\usepackage{amssymb,amsmath,amsfonts}
\usepackage{algorithmic}
\usepackage{array}
\usepackage[caption=false,font=normalsize,labelfont=sf,textfont=sf]{subfig}
\usepackage{textcomp}
\usepackage{stfloats}
\usepackage{url}
\usepackage{verbatim}
\usepackage{graphicx}
\usepackage{balance}
\usepackage{xcolor}
\usepackage{multirow}
\usepackage{threeparttable}
\usepackage{hhline}

\journal{Expert Systems with Applications}
\begin{document}
\begin{frontmatter}
\title{Cross-Domain Spatial Matching for Camera and Radar Sensor Data Fusion in Autonomous Vehicle Perception System.}

\author[agh,label2]{Daniel~Dworak}\ead{ddworak@agh.edu.pl}
\author[label2]{Mateusz~Komorkiewicz}\ead{mateusz.komorkiewicz@aptiv.com}
\author[agh,label2]{Paweł~Skruch}\ead{skruch@agh.edu.pl}
\author[agh]{Jerzy~Baranowski\corref{cor1}}\ead{jb@agh.edu.pl}

\affiliation[agh]{organization={AGH University of Kraków},
            addressline={Al. Mickiewicza 30},
            city={Kraków},
            postcode={30-059},
            country={Poland}}

\affiliation[label2]{organization={Aptiv Services},
            addressline={Ul. Podgórki tynieckie 2},
            city={Kraków},
            postcode={30-399},
            country={Poland}}
\cortext[cor1]{Corresponding author}



\begin{abstract}
In this paper, we propose a novel approach to address the problem of camera and radar sensor fusion for 3D object detection in autonomous vehicle perception systems. 
Our approach builds on recent advances in deep learning and leverages the strengths of both sensors to improve object detection performance. Precisely, we extract 2D features from camera images using a state-of-the-art deep learning architecture and then apply a novel Cross-Domain Spatial Matching (CDSM) transformation method to convert these features into 3D space. We then fuse them with extracted radar data using a complementary fusion strategy to produce a final 3D object representation.
To demonstrate the effectiveness of our approach, we evaluate it on the NuScenes dataset. We compare our approach to both single-sensor performance and current state-of-the-art fusion methods. Our results show that the proposed approach achieves superior performance over single-sensor solutions and could directly compete with other top-level fusion methods.
\end{abstract}

\begin{graphicalabstract}
\begin{figure}[!ht]
\centering
\includegraphics[width=\linewidth]{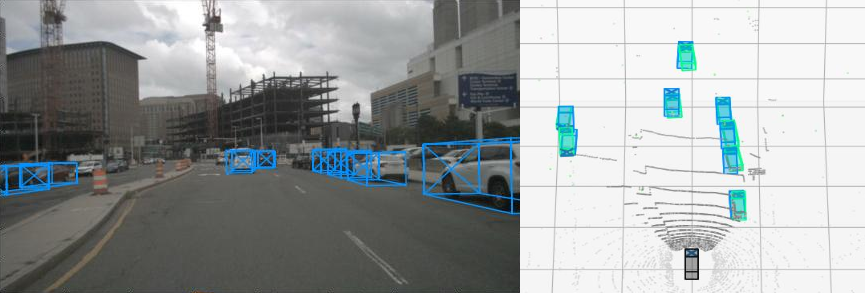}
\includegraphics[width=\linewidth]{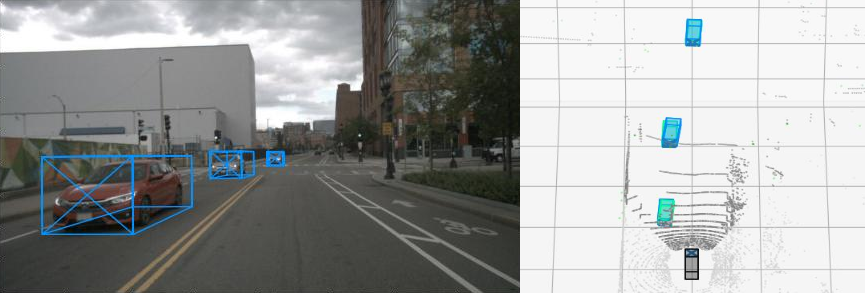}
\caption{Example results of CDSM fusion method predictions on NuScenes test data. Predicted objects are marked in blue, both in a camera and an enhanced BEV view. Green cuboids represent matched groundtruth labels. LiDAR pointcloud added for reference in BEV view.}
\label{fig:cdsm_abstract_preview}
\end{figure}
\vspace{12cm}
\end{graphicalabstract}

\begin{highlights}
\item New method of Low Level Fusion of camera and radar data within neural network structure
\item Projection-less approach based on tensor orientation matching 
\item Lightweight solution, competitive with current SOTA approaches
\vspace{18cm}
\end{highlights}


\end{frontmatter}
\section{Introduction}
Modern cars become more and more autonomous every day. Although they are still far from achieving full level 5 of autonomy \cite{TaxonomyAD}, we see significant progress in that research field. One of the major reasons for that is an advancement in artificial perception systems. In autonomous vehicles (AV), the perception system is responsible for recognising the surrounding environment: filtering out the background, detecting other road users (cars, pedestrians etc.) and important infrastructure landmarks (lane markings, traffic signs, traffic lights etc.).

To perform a perception task, the vehicle is provided with a versatile sensors suite \cite{av_sensors}. A typical configuration includes a high-resolution front camera that is used for general object detection, supplemented by lower-resolution surrounding cameras to provide a 360-degree field of view for detecting objects in close proximity to the car. Additionally, high-density LiDAR sensors are employed for precise distance measurements, while a combination of close and long-range radars is utilized to obtain accurate distance and velocity readings. 

Raw data from those sensors, in a form of an image or a poincloud, 
\begin{figure}[!ht]
\centering
\includegraphics[width=\linewidth]{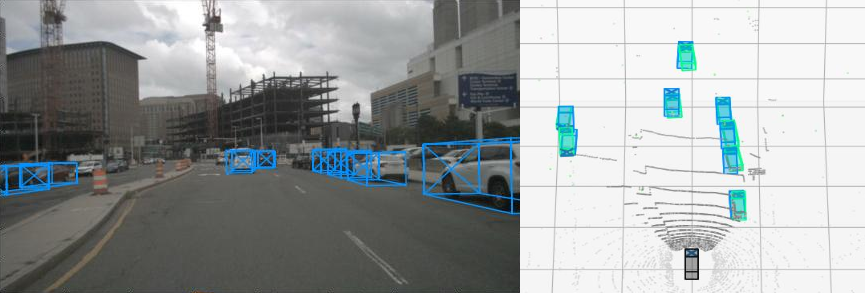}
\includegraphics[width=\linewidth]{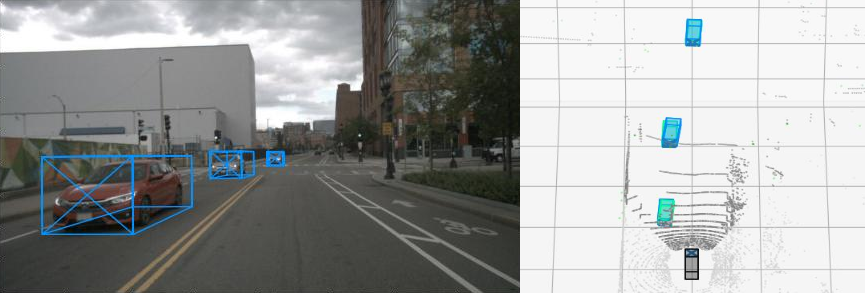}
\caption{Example results of CDSM fusion method predictions on NuScenes test data. Predicted objects are marked in blue, both in a camera and an enhanced BEV view. Green cuboids represent matched groundtruth labels. LiDAR pointcloud added for reference in BEV view.}
\label{fig:cdsm_preview}
\end{figure}
is then processed to obtain a model of an environment, used for example in path planning algorithms and safety systems. 
Creating such a model from raw sensor readings is a complicated task. It is complex to the point, where traditional algorithms could not handle the variety and the amount of data collected during different real-life road scenarios, thus machine learning techniques are used to process sensor inputs.
Especially neural networks have proven to be more than capable of performing object detection tasks. They surpass human abilities in recognizing objects in the images. Similarly, pointclouds from LiDAR and radar sensors might be difficult to interpret by humans, whereas neural networks can easily find patterns in them. 

To assure an even higher performance of the AV perception system, the fusion algorithm combines single sensor data and yields the final perception outcome. The fusion results should be more robust and benefit from each sensor's advantageous aspects \cite{fusion}. Also, in case of partial sensor blockage or other failure modes, the fusion algorithm provides an additional layer of safety. It could mitigate hazardous effects by relying on more confident sensor readings. Fusion algorithms can be classified as a high or a low level\cite{low_high}. A high-level fusion utilises information about detected objects from separate sensors and fuses them at the object level. A low-level fusion operates closer to raw input data, using information directly from each data stream. The main difference between them is that the high-level fusion operates on already processed sensor detections, whereas the low-level fusion operates on the raw or minimally processed data streams themselves. Therefore low-level fusion neural networks could find patterns in cross-sensor data, that would not be accessible at a higher (object) level.

In case of the autonomous vehicles, low-level fusion is typically done on images and pointcloud data. Images come from automotive-grade cameras and are proven to be vital for the perception system in many ways. But when it comes to pointcloud data, there are two sensors, LiDAR and radar, which may seem very similar in the output they produce, but there are major differences between them. Both sensors produce pointcloud data with accurate distance readings in 3D space, but the sparsity of LiDAR pointcloud (a few hundred thousand points) is much denser than radar (a few hundred points), hence it contains more information. That comes with a price, as rotation LiDARs are much more expensive sensors and thus are not suitable for mass production from the manufacturer's point of view. Solid-state LiDARs are cheaper, but the development of such sensors does not yet achieve the technological readiness level required for automotive-grade sensors \cite{lidar_reqs}. On the other hand, radars, which operate on radio frequencies instead of light weaves, are much more resilient to environmental effects on the road. They also provide additional information about velocities for each detected point, which could be very useful for traffic environment modelling.
Taking into account those differences, both sensors seem suitable for fusion with camera images, as they can provide complementary information. That being said, in the deep learning sensor fusion domain, there is only a handful of camera-radar fusion solutions when compare to camera-LiDAR ones.

In this article, we address the problem of low-level camera-radar fusion perception systems based on neural networks by proposing a new method of fusing data from these sensors. Based on the research presented in a related work section, we adopt a multi-view approach used for camera-LiDAR solutions, using separate single-stage architectures for camera processing and a voxelwise radar pointcloud processing. Obtained feature maps are then fused together in a novel Cross-Domain Spatial Matching (CDSM) low-level fusion block to produce an enhanced bird eye view internal representation. Based on this representation, detection heads yield 3D objects bounding boxes with related parameters. We perform experiments on the nuScenes\cite{nuscenes} dataset and show the advantages and disadvantages of our solution.

This paper is structured as follows. In section 2 we go over related work regarding single-sensor perception as well as different fusion techniques. In section 3 we present our approach to camera-radar based perception system, with detailed network architectures and the CDSM fusion method. Section 4 contains the description of conducted experiments and the results we obtained in the form of perception KPIs and improvement gain of fusion over single sensor systems respectively. Finally, we draw our conclusions in section 5.

\section{Related Work}

\subsection{Camera object detection}
An object detection task in camera images was the first field to successfully apply convolutional neural network solutions. Ever since then, researchers are constantly improving the algorithms by applying novel architectures and mechanisms to increase performance. We can divide object detection methods into two major groups: 2D  image plane and monocular 3D domain detectors.

One of the most recognizable architectures among 2D detectors is the single-shot YOLO (You Only Look Once) \cite{yolo1} network. Over time, improvements have been proposed to increase initial network performance. YOLOv2 \cite{yolo2} utilises an anchor box mechanism, where instead of predicting raw bounding boxes size, it is done relative to the most suitable, predefined anchor size. YOLOv3 \cite{yolo3} introduces multi-scale training for small, medium and large objects at different levels of neural network Feature Pyramid Network (FPN), which are then concatenated right before the Non-Max Suppression (NMS) algorithm. YOLOv4 \cite{yolo4} refine the network architecture by applying Cross-stage Partial Connections (CPS) backbone, Path Aggregation Network (PAN) \cite{pan}, attention in the form of Convolutional Block Attention Mechanism (CBAM) \cite{cbam}, CIoU \cite{diou} metric for improved loss calculation and mish activation function. Based on the same concept, RetinaNet \cite{retina} architecture introduces improvements in a form of a focal loss. The new loss function refines class imbalance problem and overall training speed and stability.

Further optimizing of single-shot detectors architecture, researchers from Google took a closer look at model scaling, namely the width, the depth and the resolution of the model. In their paper EfficientNet \cite{effnet}, through compound coefficient, they designed smaller and faster, yet better-performing architecture, which was chosen from a set of models with various width, depth and resolution parameters. Directly following that idea, the detection network architecture EfficientDet \cite{effdet} was introduced, which uses EfficientNet as a backbone. Additionally, expanding on the idea of multi-scale feature fusion, they proposed a weighted bi-directional feature pyramid network (BiFPN) to propagate internal network representation from various layers even more effectively. 
In EfficientNetV2 \cite{effnetv2}, the backbone architecture was optimized even more to achieve better training speed and parameter efficiency in terms of the model size.
Different backbone architecture is shown in Deep Layer Aggregation (DLA)\cite{DLA}. Authors propose deeper features aggregation to improve fuse information sharing across layers of the backbone. This includes both Iterative Deep Aggregation (IDA) and Hierarchical Deep Aggregation (HDA) novel features aggregation methods.

Although object detection in 3D space from a single monocular camera image is a much more complicated task, recent studies show, that specific neural network architectures are also capable of achieving meaningful results. In CenterNet \cite{centernet3d}, the extension of 2D model \cite{centernet}, the proposed approach is to divide 3D object detection into two steps. The first is an anchorless prediction of the centre for a given cuboid in the image and the second step is the regression of all 3D parameters such as depth, 3D dimensions, and rotation angles. After a projection of the predicted centre, 3D results are obtained. Similarly, in FCOS3D \cite{fcos3d}, authors use a set of predefined landmark 3D points in the image to perform 2D centreness prediction and based on 2D position and depth project it to 2.5D space. The rest of the parameters are regressed in 3D space as well, to yield final object predictions.
  
\subsection{Pointcloud object detection} 
An input sensor data from LiDAR and radar comes in a form of a list of points coordinates and corresponding features. Depending on the sensor, the features can be reflection intensity for LiDAR or cross-section and velocities for radar. The list of points is often referred to as a pointcloud. Pointcloud processing poses some challenges with regard to neural networks. The network must be invariant to all permutations of input points; changing the order of the data in the input list should yield the same results. The length of such a list could also vary depending on the sensor reading, but neural network architecture, due to how it is structured, tends to except fixed input size.
On the other hand, when scattered in 3-dimensional space, pointcloud data is very sparse (95\% of that space remains empty). Those issues resulted in two approaches to processing pointclouds with neural networks: pointwise and voxelwise. 

A pointwise approach in classification network Point-Net \cite{pointnet} uses transformation dense layers to extract features for each point individually, although the weights of those layers are shared. To assure invariance to points order, the max pooling layer is used to extract global features. Another pointwise architecture, PointRCNN \cite{pointrcnn} takes a two-stage approach for 3D object detection tasks. Stage one, segments points from the background and generates a small number of detections in a bottom-up manner. In stage two, those detections are refined pointwise with respect to local spatial features and global semantic features, which results in accurate bounding boxes and confidence scores.

For a detection problem, voxelwise methods are more commonly used. Introduced in VoxelNet \cite{voxelnet}, the idea is to scatter points in 3D space, to minimize computational effort and data sparsity, whole space is divided into smaller cuboids called voxels. Each voxel has features calculated by Voxel Feature Extractor (VFE) layers based on points inside it. After feature extraction, a fixed-size output tensor is processed by 3D convolutional layers to produce 3D detections. Pointpillars \cite{pointpillars} algorithm 
changes feature extraction by stacking voxels vertically (along the z-axis) into pillars. By doing so, the output of the extraction is a three-dimensional tensor rather than a four-dimensional one, like in VoxelNet. This enables the usage of 2D convolutions instead of 3D ones, and as shown in the paper tremendously increases inference time (up to real-time).  

In a recent work called PV-RCNN \cite{pv-rcnn}, authors propose both pointwise and voxelwise processing methods combined into one network architecture. In addition to normal voxel feature extraction, feature maps from voxelwise subnetworks are fused with pointwise features in the original Voxel Set Abstraction Module. As a result of the fusion, Keypoint Features are produced and passed to the detection head to amplify certain regions in the output grid.

Radar-only 3D object detection, on the other hand, is a less popular subject of research. In recent NVRadarNet \cite{nvradarnet}, authors use sensor peak detections, rather than raw antennas signal, to create a sparse radar pointcloud. Such pointcloud is scattered onto a BEV grid and processed via an encoder-decoder model to yield 3D objects. They present the results on the NuScenes dataset, but they are far from what LiDAR sensors achieve in terms of KPI metrics for 3D object detection tasks.

\subsection{Fusion architectures}
Sensor fusion algorithms fuse the data from different sensors to obtain improved performance. This is especially true for image and pointcloud data, as cameras and LiDARs or radars perceive the environment in a completely different, but complementary manner. Due to significant differences in sensor readings domains, image camera view and pointcloud 3D surrounding view, the fusion poses a problem of incorporating those two sources of information together.

In multi-view setups (AVOD \cite{avod}, MV3D \cite{mv3d}, PointFusion \cite{pointfusion}), each sensor input is processed by a distinct subnetwork to obtain view-specific feature maps. Those views are typically a Bird's Eye pointcloud View (BEV), a front pointcloud view (3D points projected to camera plain view) and a camera view. Based on concatenated feature maps, the fusion region proposal network determines Regions of Interest (ROIs) for detection heads. In those solutions, the fusion process is typically performed in an end-to-end manner, where the specific method of merging detailed information is determined by the distribution of network weights learned during the training process. 

A different approach to fusion is shown in PointPillars++ \cite{pointpillars++} and Joint 3D Object Detection \cite{joint3d}. Authors enhance pointcloud data in LiDAR front view by incorporating camera pixel information into corresponding points. The fused front view is then processed as a pointcloud with additional features by a neural network. By performing this deterministic step during input processing, a certain fusion method is forced upon the neural network architecture and the usage of both information sources is better conditioned.

Novel solutions for the drivable area and road detection field can be also applied to object detection fusion. In \cite{roadfusion1}, besides projecting LiDAR pointcloud onto a camera image, authors apply a depth-completion algorithm to create dense depth maps from sparse points in a front view, thus providing more information to the fusion algorithm. The opposite solution was proposed in \cite{roadfusion2}. Instead of projecting points onto the image, the pixels data was projected onto a pointcloud BEV occupancy grid, resulting in a completely different fusion domain, but achieving comparable results.

All the fusion methods discussed were designed with respect to pointcloud data from a LiDAR sensor. There are only a handful of fusion deep learning solutions that utilize camera images and a radar pointcloud, as mentioned in \cite{camrad-3Dod}. In the same paper, authors propose a base multi-view network architecture and present their results, which are however inferior in terms of performance to state-of-the-art camera-LiDAR fusion methods. They explain possible reasons for this gap to be a matter of a small training dataset, as well as pointcloud data differences specific for both sensors. On the other hand, CRF-Net \cite{crfnet} achieves satisfying results for a camera-radar fusion on a newer, much larger NuScenes dataset \cite{nuscenes}. The approach for the fusion is to enhance a camera image with projected radar points but in the form of vertical lines in the image. Authors show improvements over a baseline camera-only object detection network. Regardless, the objects are detected in a 2D camera image space, rather than a 3D domain. 

Predictions in 3D space are obtained in recent 
CenterFusion \cite{centerfusion} architecture. The fusion is done on a camera image processed similarly to the CenterNet vision-only model approach but with additional information from radar detections. First, a 2D centre point and object features are predicted in the image, which is then associated with extracted radar features via the frustum association mechanism. The fusion of two sensor feature maps leads to final 3D predictions.
In FUTR3D \cite{futr3d}, authors proposed a framework to fuse camera images with both LiDAR and radar pointclouds. They employ a query-based modality agnostic feature sampler to fuse all sensor features and accommodate a transformer decoder to predict 3D objects directly. 

Those fusion solutions are currently achieving top positions in the official NuScenes 3D object detection ranking \cite{nuscenes_ranking}. In the next section, we present our approach to camera and radar sensor data fusion. We target the full 3D prediction domain, as such predictions are more desirable input for perception systems, although they are also more challenging to obtain. To that end, we propose a new simple yet effective method to fuse image and pointcloud feature maps to accomplish said task.

\section{Proposed Approach}

\begin{figure}[!ht]
\centering
\includegraphics[width=\linewidth]{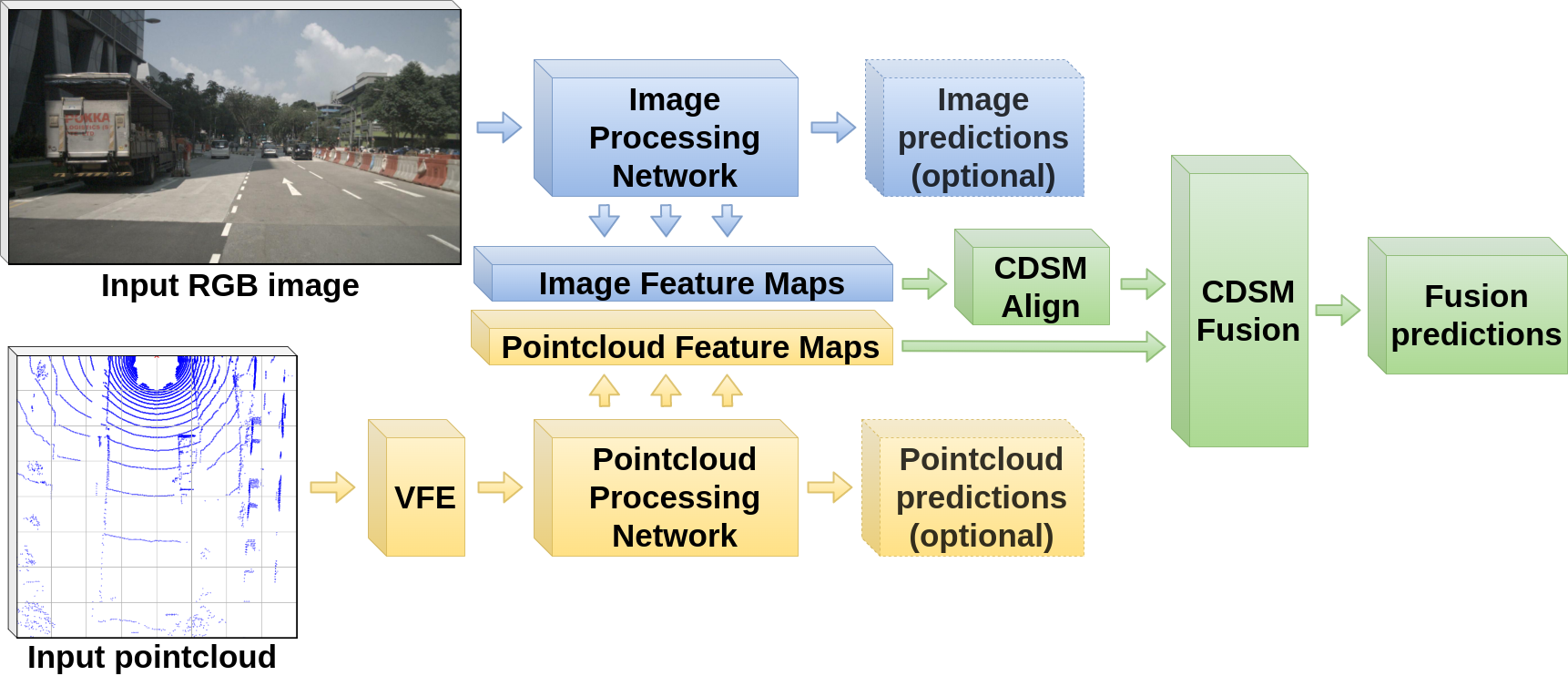}
\caption{Whole solution pipeline with camera image and pointcloud list inputs, image processing network in blue, pointcloud processing network in yellow, both with optional outputs and CDSM fusion in green with main fusion predictions output.}
\label{fig:all_graph}
\end{figure}

Our approach to fusion adopts a multi-view setup concept. We use separate network architectures to process both camera images and radar pointcloud data (Figure \ref{fig:all_graph}). Input from the camera is processed in a 2D image domain, whereas radar pointcloud is processed in a 3D space in an enhanced BEV. Both neural networks could produce their respective output, predictions in related domains. Additionally, for the purpose of low-level sensor fusion, we also introduce novel Cross-Domain Spatial Matching (CDSM) fusion block. Our goal is to fuse feature maps from intermediate networks layers to create a single fusion output in a 3D space. The main issue with those feature maps is that they come from completely different domains (2D camera and 3D BEV), thus in order to benefit from both sources we need to spatially align them before the fusion, which is done in the CDSM block.

\subsection{Image network architecture:}
\begin{figure}[!ht]
\centering
\includegraphics[width=\linewidth]{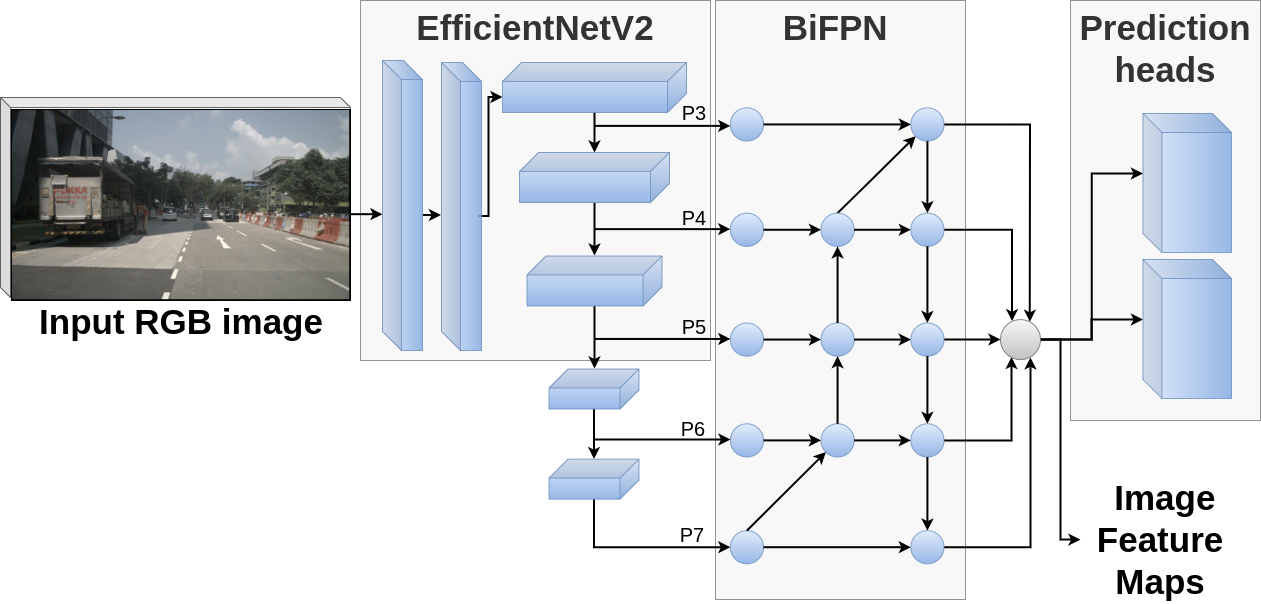}
\caption{Camera network architecture.}
\label{fig:cam_graph}
\end{figure}

For camera image processing we designed a single-stage detector based on EfficientDet network structure. There are 3 main elements that constitute our model (Figure \ref{fig:cam_graph}): an EfficientNetV2 backbone for initial features extraction, a BiFPN that aggregates and merges features across different levels of abstraction and finally classification and regression heads, which predict the final outcome. 

While the core concept remains unchanged, we made modifications to the network structure in order to optimize it for our specific purpose. An input resolution was changed to 512x384 pixels to better suit the NuScenes dataset image aspect ratio. We extract features from the backbone network at 3 stages corresponding to official P3-P5 levels (1/8, 1/16 and 1/32 of an input size). Then, we artificially added levels P6 and P7 to match the required input of the BiFPN block. After 4 repeats of BiFPN, refined features are passed to the classification head for object class prediction and score, as well as the regression head, for bounding boxes coordinates and sizes.

We used ImageNet pretrained weights for the EfficientNetV2 backbone and randomly initialized ones for the BiFPN and prediction heads. Through experiments, we decided to choose a mix of LeakyRelu and Mish activation functions across all layers. We also tried different normalization layers: BatchNorm, GroupNorm, InstanceNorm and LayerNorm, where the last one proves to work best for us.

Finally, the model predicts objects on 5 different scales (output grid size 1/8, 1/16, 1/32, 1/64 and 1/128 of an input size) with respect to corresponding anchors. Anchors have been automatically generated based on each grid size and combinations of 3 scale factors and 3 ratio factors, which resulted in 9 anchors per grid cell.
In order to yield final results, we use the Non-Max Suppression algorithm on detections from all 5 scales simultaneously to remove duplicates and overlapping detections.  

\subsection{Pointcloud network architecture:}
For the radar pointcloud processing network, we took inspiration from architectures created to process LiDAR pointclouds. We process radar data with a voxel-wise approach, we divided the whole 3D space into a voxel grid of size 1m x 1m x 1m, due to high data sparsity. Then, in Voxel Feature Extractor (VFE), based on radar points in each voxel, we calculate its features. Maximum points per voxel are limited to 5, as VFE requires a fixed amount of points. To ensure sufficient inference time, we stack voxels along Z-axis to transform voxel feature tensors from 4D to 3D. 

\begin{figure}[!ht]
\centering
\includegraphics[width=\linewidth]{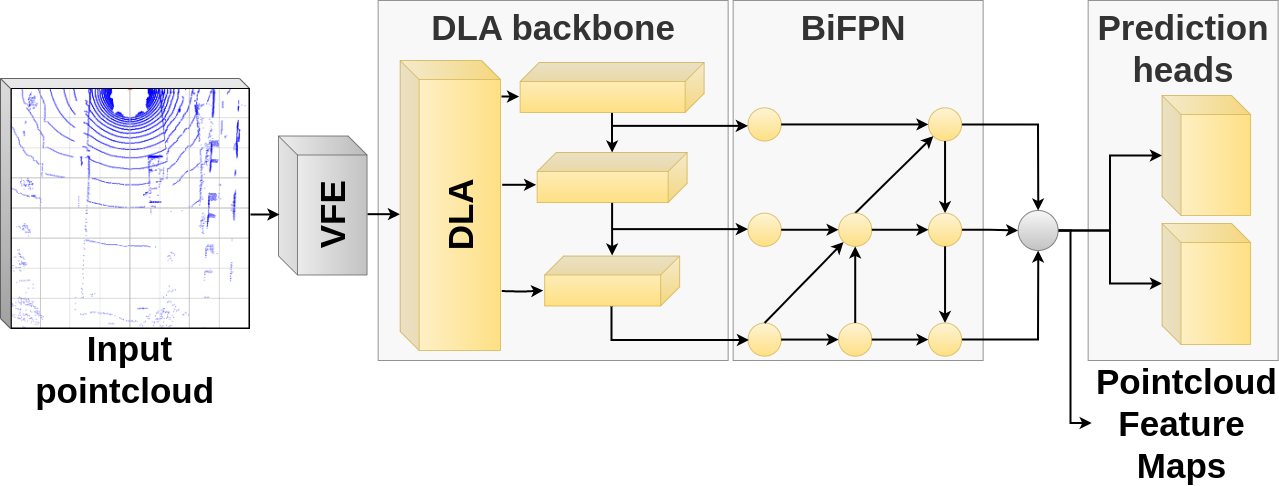}
\caption{Radar network architecture.}
\label{fig:pc_graph}
\end{figure}

After Voxel features extraction, the pointcloud network architecture is similar to the previously described image network: it has a backbone, a BiFPN block and prediction heads (Figure \ref{fig:pc_graph}). Because we are no longer using ImageNet pretrained weights, we changed the backbone to DLA34, which we further significantly modified for a pointcloud processing purpose. Our new backbone is much smaller than EfficienNet but still provides aggregation functions of a DLA architecture. The BiFPN blocks and prediction heads have also been reduced in terms of the number of layers. The reason for this is the sparse nature of radar pointcloud and a rather low amount of information to process (compare to the camera or the LiDAR). 

The output of the pointcloud processing network is a set of 3 BEV grids of sizes 80x80, 40x40 and 20x20 (cell size of 1m, 2m and 4m respectively), that covers 80x80m ROI. The objects are also predicted with respect to auto-generated anchors per each scale. The difference is in encoding an additional Z dimension of the centre and height of a bounding box, as well as yaw rotation angle in 3D for each prediction. Combining predictions in an NMS algorithm yields final results.

\subsection{CDSM fusion:}
The main innovation proposed in our solution is a fusion block called Cross Domain Spatial Matching (CDSM). The core concept of this fusion block is based on the spatial alignment of the information contained in sensor readings from the camera image and radar pointcloud, as respective feature maps from each network intermediate layer are initially misaligned. CDSM  consists of 2 major elements: Domain Alignment and Aligned Features Fusion.

\begin{figure}[!ht]
\centering
\includegraphics[width=\linewidth]{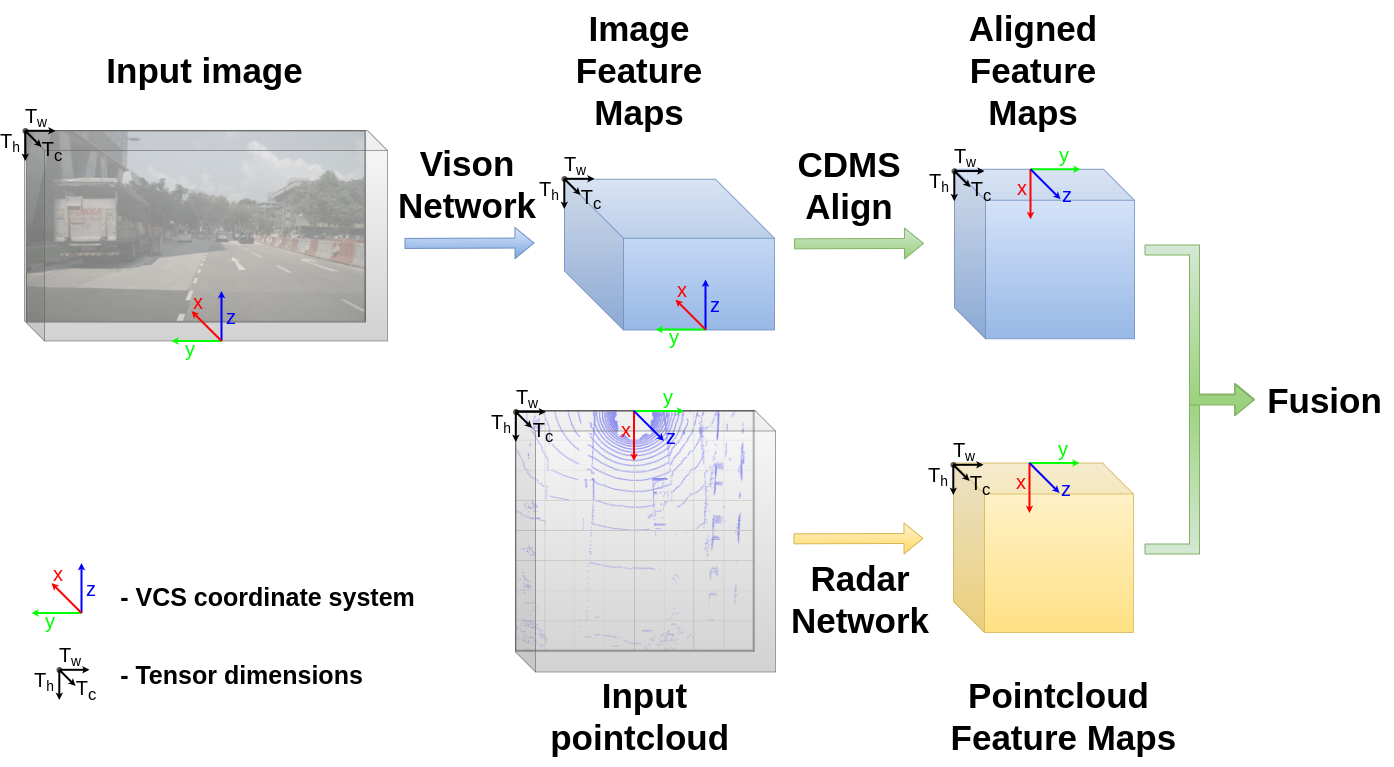}
\caption{CDSM Domain alignment diagram. Image features are marked in blue, pointcloud features in yellow and the
fusion elements in green. As we can observe, the initial feature maps orientation in the VCS coordinate system is mismatched. We apply a custom CDMS align layer, rotating 2D image features to match those of the pointcloud, in preparation for the fusion architecture.}
\label{fig:cdsm_align}
\end{figure}

To better understand this idea, we first introduce a vehicle coordinate system (VCS). The VCS is centred on the car's front axle, with X-axis pointing forward, Y-axis pointing to the left of the car and the Z-axis straight up. Considering the VCS, we can position sensor readings, namely the image and the pointcloud voxel grid in this one, unified space. 
As shown in Figure \ref{fig:cdsm_align}, related 3D tensors for both inputs have different orientations. For a camera image, the first 2 dimensions correspond to VCS ZY-plane and learned features (initially RGB values) span throughout X-axis. In the case of a pointcloud voxel grid, the first 2 dimensions correspond to VCS XY-plain and features (initially stacked VFE outputs) span along Z-axis.
The latter representation is consistent with the expected single-shot perception network output, that is a BEV grid (in XY-plain) with detected objects and its parameters. However, fusing information from the camera poses a problem, as those tensors contain features from different perspectives.
In the CDSM fusion block, we address this fusion problem with the following solution. 

\textbf{\textit{Domain alignment.}} Before fusing information from both views, we align the tensors to match their spatial orientation in VCS. To do so, we implement a custom CDSM rotation layer to perform such an operation. In principle, we used a chain of quaternion rotations to calculate the final rotation matrix and apply it (via matrix multiplication) to tensor indexes. We also shift the new indexes by calculated offset to align the (0,0,0) tensor index (as some rotations result in negative index values). Finally, we gather all values from old indexes and scatter them across rotated output tensor according to new indexes. We use the CDSM rotation layer to match camera and radar feature maps tensors with respect to the VCS. The parameters and order of camera features tensor rotations are as follows: firstly rotating by 180 degrees around the first dimension (VCS Z-axis) and then by 90 degrees around the second dimension (VCS Y-axis). 
It is worth mentioning that chosen combination and order of rotate operations not only assures that both tensors are in the same orientation with respect to VCS but also both centres of the VCS are aligned in the same position. Such alignment can not be achieved with any combination of permutations and/or transpositions of the input tensor dimensions.

\begin{figure}[!ht]
\centering
\includegraphics[width=\linewidth]{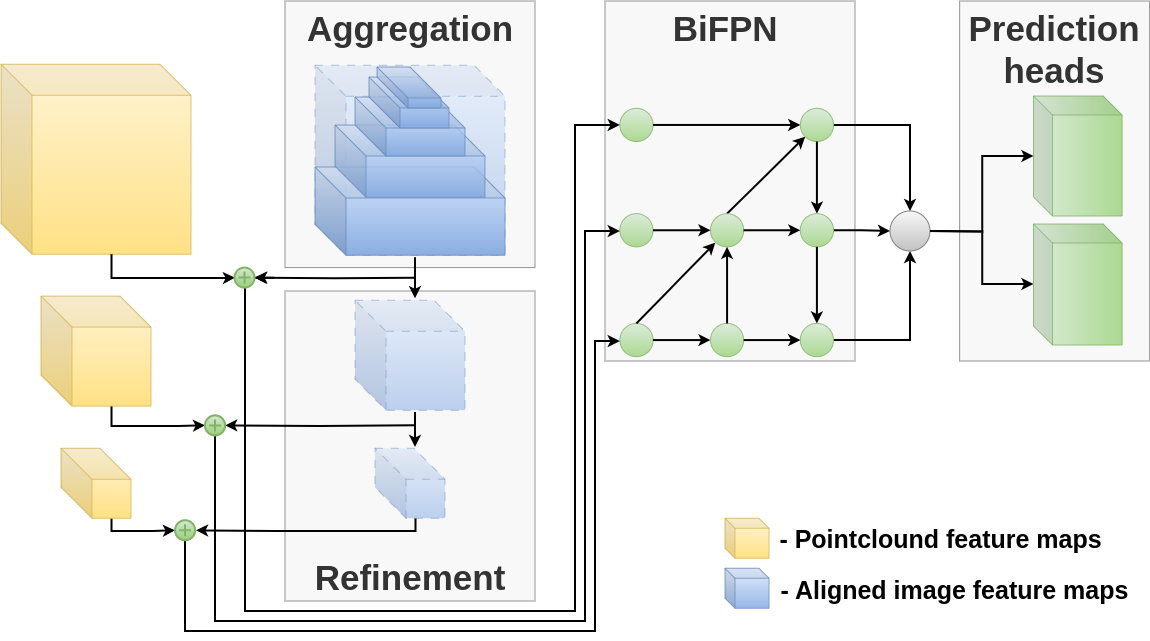}
\caption{CDSM fusion architecture. Aligned image features are marked in blue, pointcloud features in yellow and the
fusion elements in green. We apply aggregation and refinement steps to image feature maps and fuse them together with pointcloud data using tensor concatenation. Then, after a BiFPN processing, prediction heads output final fused 3D predictions.}
\label{fig:cdsm_fusion}
\end{figure}

\textbf{\textit{Aligned features fusion.}} With both tensors spatially aligned we are able to merge information from both views in a fusion block. Our proposed CDSM fusion method (Figure \ref{fig:cdsm_fusion}) can be divided into three following stages. 

At first, we take camera feature maps from different scale levels and we aggregate them on a single BEV map. The reason to do so is that those feature maps are responsible for detecting objects of different sizes in a camera plane, and thus due to perspective mapping, they correspond to particular regions in the BEV domain at a certain distance from the camera sensor. We used Grad-CAM\cite{gradcam} visualization method to determine the distance ranges for features from each scale level. Additionally, when aggregating feature maps, we take into account the relation of the camera sensor field of view with respect to the output BEV grid. This assures that the features are not placed in a 3D space not visible in the image.

After the aggregation, we propose a features refinement step. It consists of several 2D convolutional layers in the BEV domain. Similarly to the backbone concept, we process the features from detailed to more general ones, creating smaller grid representations of the same BEV area. This step allows us to obtain the relations between different feature maps throughout the training process in an end-to-end manner, instead of manually invoking them onto the model. It also creates higher-level features, that capture larger areas in a BEV. Finally, the result of the refinement is a set of 3 different BEV grid feature maps from the camera sensor in a 3D domain, which we could directly fuse with pointcloud feature maps.

The fusion of camera and pointcloud features is a relatively simple task now, as we converted them into the same coordinate system. Aggregated and refined camera feature maps in BEV correspond spatially to those obtained during radar data processing. During architecture design we assure compatible grid sizes so that we could concatenate both grid tensors together, stacking camera and radar features for each grid cell along the channel dimension.
We apply another BiFPN block to concatenated feature maps at different levels in order to further combine both sensors' information into a single 3D internal representation. This representation is used in prediction heads to yield final 3D object predictions.

\section{Experiments \& Results}

\subsection{Dataset}
\begin{figure}[!ht]
\centering
\includegraphics[width=0.95\linewidth]{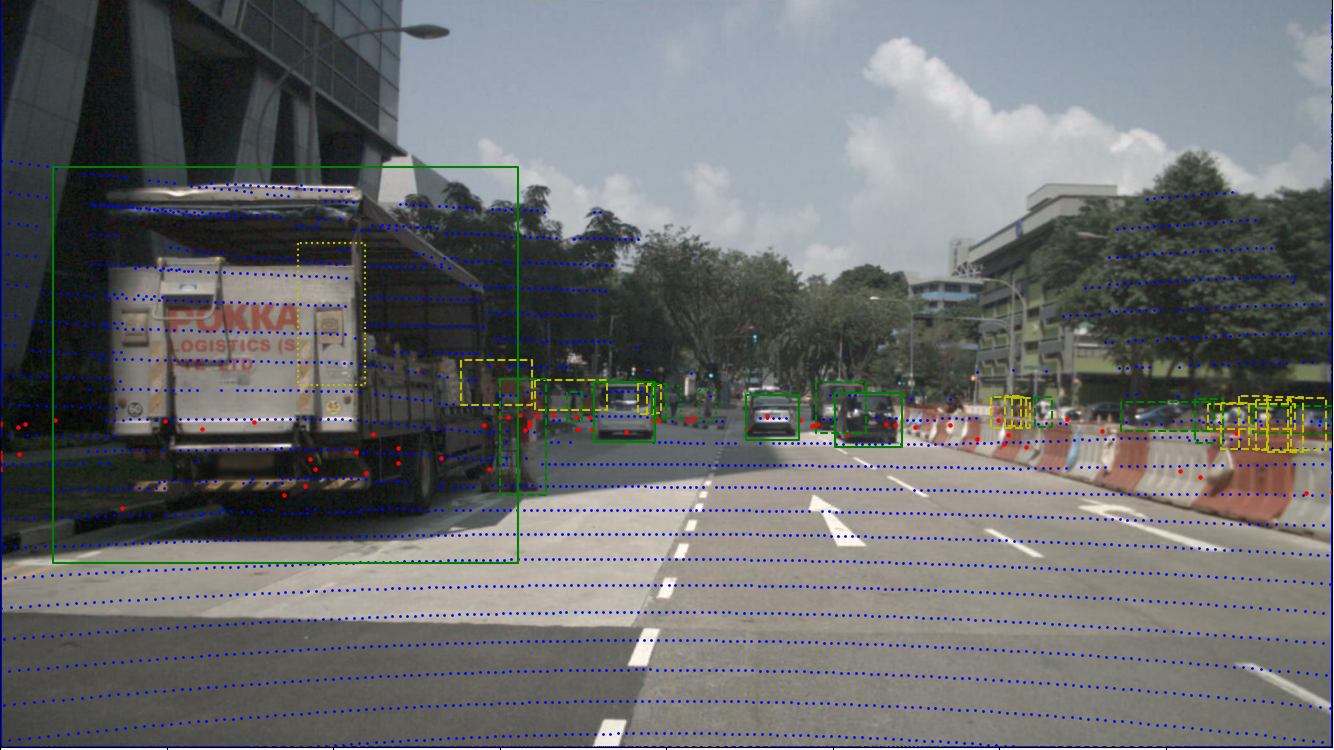}
\caption{NuScenes camera view with projected LiDAR points (blue), radar points (red) and labels. Labels with visibility over 50\% are marked in green, otherwise in yellow. Additional overlays are shown to present data, camera processing network is fed with raw RGB image. Best viewed in colour.}
\label{fig:nu_img}
\end{figure}

\begin{figure}[!ht]
\centering
\includegraphics[width=0.95\linewidth]{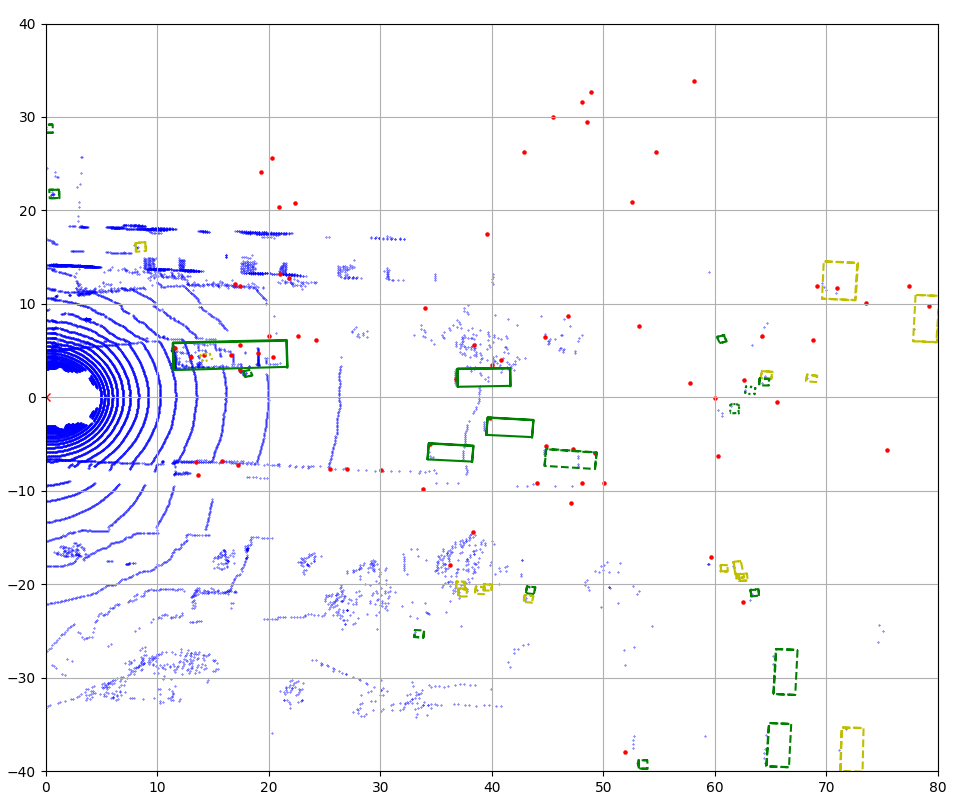}
\caption{NuScenes Bird's Eye View with projected LiDAR points (blue), radar points (red) and labels. In addition to colour-coded label visibility, solid line style indicates both LiDAR and radar points within the labelled object, dashed line only LiDAR points and dotted line neither LiDAR nor radar points within the labelled object. Best viewed in colour.}
\label{fig:nu_bev}
\end{figure}

We train our fusion solution on a popular automotive dataset called NuScenes, published in 2019. Recorded scenes in this dataset come from real-world test drives across different environments and cities. Detailed information about the dataset is presented in \cite{nuscenes} paper. We used NuScenes version 1.0.
Regarding sensors setup, the car was equipped with 6 cameras, 1 top LiDAR and 5 radar sensors. For the purpose of this research, we only use a front view RGB camera along with LiDAR and radar readings (Figure \ref{fig:nu_img} and \ref{fig:nu_bev}) within the chosen field of view (FOV). As a FOV we decided to take into account only the area where pointcloud data and camera view overlap, in the previously mentioned VCS this area is bounded from 0m to 80m in the X-axis (in front of a car), from -40m to 40m in Y-axis (from right to left) and from 0m to 5m in Z-axis (the height). Dataset split follows common train, validation and test sets division and the sizes of those are 19872, 8111 and 4485 samples respectively. 

\textbf{\textit{Image data preprocessing.}} NuScenes front RGB camera has a resolution of 1600x900 pixels. This is quite a large size to be processed by the neural network. To mitigate computing requirements for our model we decide to resize images to 512x384 pixels resolution. We also used a letterbox resizing mechanism to keep the image aspect ratio and normalization of pixel values from 0 to 1 across all RGB channels.

\textbf{\textit{Pointcloud data preprocessing.}} Pointcloud data comes in the form of a list of points with XYZ coordinates in the sensor coordinate system along with sensors' specific readings, intensity for LiDAR and parameters like velocities, cross-section etc. for radar. Firstly we map those coordinates to the defined VCS. Additionally, we remove points that fall outside our defined FOV, as they pose no useful information to the fusion algorithm. Meanwhile, this step helps with the network inference speed, as fewer data points need to be processed. Clipping pointclouds to the FOV results in fewer points from LiDAR and radar sensors for further processing. The average number of points in pointcloud per sample is 13567 for LiDAR and 45 for radar data.

\begin{table*}[!ht]
\renewcommand{\arraystretch}{0.7}
\caption{Labels detailed information regarding label visibility for the camera, LiDAR and radar sensors across the whole dataset. Statistics were captured using NuScenes human annotation information per each labelled object present in the data.}
\label{tab:labels_info}
\centering
\begin{tabular}{|c|c|c|c|}
\hline
 & All classes & Cars & Pedestrians \\ 
\hhline{=|=|=|=}
Total labels count & 549289 & 219328 & 116952 \\ \hline
Camera visibility over 40\% & 346475 (63\%) & 126355 (58\%) & 72459 (62\%) \\ \hline
Labels with LiDAR points & 451621 (82\%) & 170519 (78\%) & 102295 (87\%) \\ \hline
Labels with radar points & 173836 (32\%) & 101049 (46\%) & 12839 (11\%) \\ \hline
Mean LiDAR points per label & 97 & 127 & 14 \\ \hline
Mean radar points per label & 2.26 & 1.96 & 1.14 \\ \hline
\end{tabular}
\end{table*}

\textbf{\textit{Label data preprocessing.}} NuScenes dataset was annotated manually by humans in 3D space, based on LiDAR pointcloud and camera images. Labels are divided into classes like cars, pedestrians, trucks etc. Each class has sub-classes i.e. sitting pedestrian, walking pedestrian etc. For the purpose of object detection, we only distinguish top-level classes. 
In the camera object detection, we transform corners of 3D labels onto the camera image plane and draw the smallest rectangle bounding box containing all projected points. We also scale those bounding boxes according to the original image resize coefficients. For pointcloud and fusion detection, we took labels straight from the NuScenes database, as they are placed in the same space, but the following postprocessing was done regarding label filtering. 

NuScenes labels provide additional information about the visibility of the objects in the camera image, as well as a number of LiDAR and radar points that belong to a given labelled object. This information allowed us to filter some of the labels, as there were no data required to detect those objects in this particular sensor setup. Based on characteristics from Table \ref{tab:labels_info}, we decided to use only labels with visibility over 40\% as camera object detection groundtruth and labels that have at least one radar detection as 3D enhanced BEV object detection groundtruth. For the fusion we want to prove its robustness, thus groundtruth should be either visible in the camera or has radar detection or both. 
Lastly, we decided to focus on car objects solely, as the radar detections for that class are reliable enough and true benefits of fusion with that sensor could be observed. 

\subsection{Training}

In order to prove our method and show fusion benefits over single-sensor solutions, we trained both camera and radar detection networks separately, as well as combined a multi-sensor fusion model with the CDSM block.  

Starting from single sensor architectures, we trained camera 2D and radar 3D processing models. Apart from ImageNet pretrained weights for the camera EfficientNetV2 backbone, we used the random Xavier initialization method for the DLA backbone, BiFPNs and prediction heads. For classification heads, we utilized focal loss with fine-tuned hyperparameters $\alpha=0.25$ and $\gamma=1.5$ and weighted mean square error loss for regression heads across both models. The optimization process was done with Adam, the initial learning rate of $lr=3e^{-5}$ and the cosine annealing learning rate scheduler for runtime adjustments. We trained models until early stopping, monitoring validation loss, and did not show any improvements in 5 consecutive epochs. 

Both trained models achieve decent results, making them suitable sub-models for the CDSM fusion. However, we find it rather unrelated to compare 2D camera metrics with 3D radar and fusion ones. To that end, we trained another vision-only model, that predicts objects in 3D space based on monocular camera images. It was trained on top of the existing 2D model, but after obtaining 2D feature maps from BiFPN we applied CDSM alignment and aggregation layers, without any fusion with radar data. Such transformation of 2D features into 3D space enables direct prediction of objects in that domain, similar to the CDSM fusion concept.  

Lastly, we took both single-sensor models and conducted end-to-end CDSM fusion model training. We used previously pretrained sub-models to obtain sensor-specific feature maps from camera and radar data and apply CDSM alignment, aggregation and fusion to them. Training hyperparameters were similar to single-sensor ones. We also experiment with fine-tuning the pretrained networks. At first, we freeze them and trained only the fusion part of the architecture. Afterwards, we optimized them as well during the training, making adjustments precisely for the fusion purpose.

\subsection{Results}

Evaluation of obtained results was done on part of the NuScenes dataset, which consists of particular scene sequences not used in the training and validation process. We utilized the most popular performance metric for object detection tasks called mean average precision score (mAP), which is based upon precision-recall relation at different threshold levels. Moreover, mAP is highly dependent on the true positive association method, thus we explicitly state what method we used in each experiment, whether it is intersection over union (IoU) or absolute distance (DIST) between centres of 3D cuboids. 

\begin{figure}[!ht]
\centering
\includegraphics[width=\linewidth]{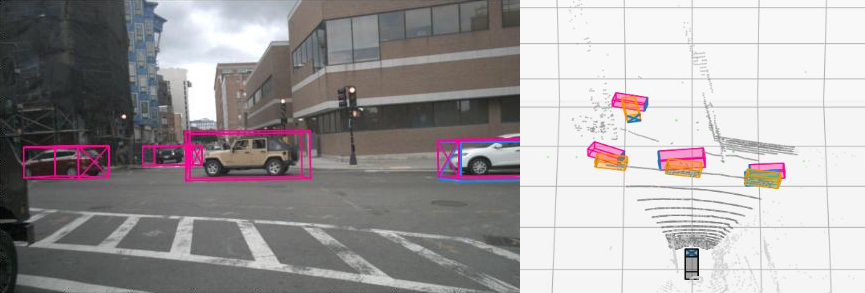}
\includegraphics[width=\linewidth]{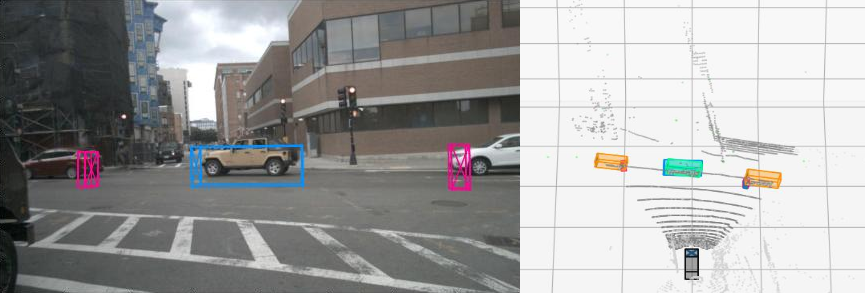}
\includegraphics[width=\linewidth]{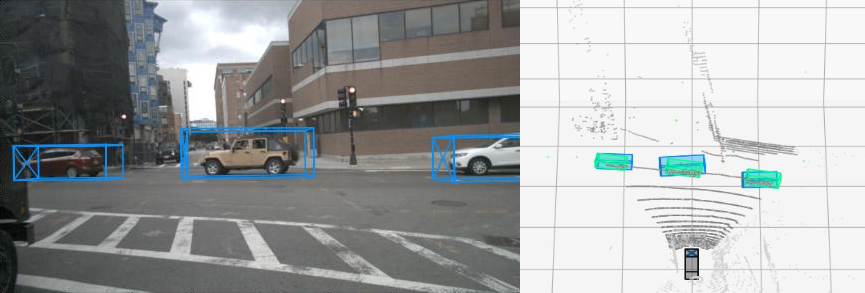}
\caption{Prediction results for the same test dataset scene for camera-only, radar-only and CDMS fusion models respectively from top to bottom. On both. images and corresponding 3D views, we marked predicted bounding boxes matched as true positives (predictions in blue, matched targets in green), false detections (magenta) and missed objects (yellow). LiDAR pointcloud added for reference in BEV view.} 
\label{fig:cdsm_results}
\end{figure}

In figure \ref{fig:cdsm_results}, we present results for the same dataset sample obtained from a single camera and radar sensor networks, as well as from the fusion one. For the camera-only model, we can observe a high object detection rate, as well as accurate general size estimation. On the other hand, depth distance prediction in 3D is rather inaccurate, which results in mismatched detections due to failed association process. The radar-only model, on the contrary, predicts accurate positions, but due to the low amount of detections, it struggles to forecast objects' dimensions and orientation properly. Finally, the fusion model makes use of both sensors' advantages and mitigates their weaknesses. The fusion of radar data precise position readings and camera ability to predict the accurate size, orientation and class results in the CDSM model vastly outperforms single sensor models.

In Table \ref{tab:kpi_ours}, we arranged the results metrics for trained single sensor and fusion models, along with modality, predictions domain and association method used to calculate the mAP score.

\begin{table}[!ht]
\caption{mAP performance metric comparison for all of our single sensor models as well as fusion ones. Modality corresponds to C - camera, and R - radar sensors. The detection domain is either a 2D image space or a 3D-enhanced BEV grid. The association method for the image network is intersection over union with a 0.2 threshold value (IOU20), whereas 3D predictions were matched with 2m distance criteria (DIST2).}
\label{tab:kpi_ours}
\centering
\begin{tabular}{|c|c|c|c|}
\hline
Method & Modality & Association & mAP$_{car}$ \\ 
\hhline{=|=|=|=}
Vision model & C (2D) & IOU20 & 0.741 \\ 
\hhline{=|=|=|=}
Vision model & C (3D) & \multirow{4}{*}{DIST2} & 0.461 \\ 
\cline{1-2}\cline{4-4}
Pointcloud model & R (3D) &  & 0.324 \\ 
\cline{1-2}\cline{4-4}
CDSM Fusion & C+R (3D) &  & 0.523 \\ 
\cline{1-2}\cline{4-4}
CDSM Fusion (FT) & C+R (3D) &  & 0.681 \\ \hline
\end{tabular}
\end{table}

Although the vision-only model in 2D has the highest mAP score, it is the only solution that yields predictions in a 2D image space, which is a far less complex task than a 3D object detection. When we consider the vision-only model, but in a 3D domain, the mAP score is significantly lower, as a result of an additional depth estimation task for each object, based on a single image frame only. The radar pointcloud-based model score is even lower, due to discussed high radar detections sparsity issue. The predictions, even if present, are often considered false positives, as the association conditions are not met, mostly because of poor size estimation.

The fusion model outperforms both single sensors by a large margin. Considering the same association metric, the mAP is much higher, which indicates that more objects are detected correctly with better overall accuracy. On top of that, a fine-tuned version, where we do not freeze single sensor submodels and adjust their parameters during the
\begin{figure}[!ht]
\centering
\includegraphics[width=\linewidth]{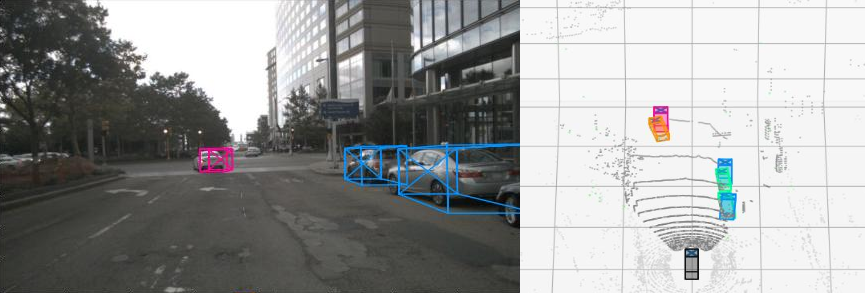}
\includegraphics[width=\linewidth]{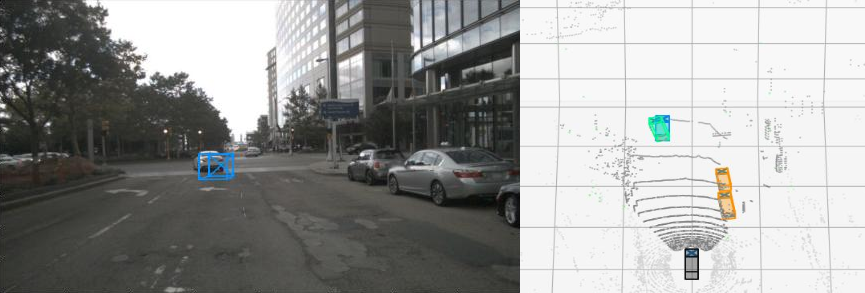}
\includegraphics[width=\linewidth]{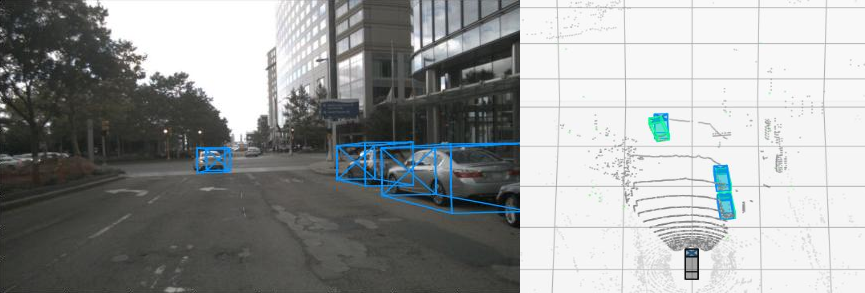}
\caption{Corner case example for camera, radar and fusion models respectively from top to bottom. Same bounding box color coding as in Figure \ref{fig:cdsm_results}. Even when single-sensor models fail to predict all objects in the scene, a fusion model utilizes both sensors' data to improve upon each of them.}
\label{fig:cdsm_corner}
\end{figure}
training, achieves even better results, as the internal representation of camera and radar data is accommodated for fusion purposes.

Another situations, where fusion benefits could be observed, are single sensor failures to detect particular objects. In Figure \ref{fig:cdsm_corner} we showcase such a corner case, in which the camera model positively identified parked cars on the right, but the proceeding vehicle is detected too far from the groundtruth position. On the other hand, the radar model detection of the car in front is really precise, while parked cars are missed completely. The fusion model predicts all objects and even improves on both sensors' position and size estimation.

Finally, we compare our results to other state-of-the-art solutions in Table \ref{tab:kpi_sota}. In order to do so, we calculated mAP score according to the official NuScenes ranking, which is a mean of mAP for four different association methods, namely DIST 0.5m, 1m, 2m and 4m.

\begin{table}[!ht]
\renewcommand{\arraystretch}{0.7}
\caption{Comparison of different state-of-the-art solutions to our approach using mAP performance metric. Modality corresponds to C - camera, L - LiDAR and R - radar sensors. The detection domain is either a 2D image space or a 3D-enhanced BEV grid. We used the official NuScenes association method.}
\label{tab:kpi_sota}
\centering
\begin{threeparttable}
\begin{tabular}{|c|c|c|c|}
\hline
Method & Modality & Association & mAP$_{car}$ \\ 
\hhline{=|=|=|=}
PointPillars & L (3D) & \multirow{6}{*}{\vtop{\hbox{\strut Average mAP}\hbox{\strut over all DIST}\hbox{\strut 0.5,1,2 and 4}}} & 0.684 \\ \cline{1-2}\cline{4-4}
FCOS3D & C (3D) &  & 0.524 \\
\hhline{=|=|~|=}
CRFNet & C+R (2D) &  & 0.559 \\ \cline{1-2}\cline{4-4}
CenterFusion & C+R (3D) &  & 0.509 \\ \cline{1-2}\cline{4-4}
FUTR3D & C+R (3D) &  & 0.52-0.54* \\ \cline{1-2}\cline{4-4}
CDSM (Ours) & C+R (3D) &  & 0.535 \\ \hline
\end{tabular}
\begin{tablenotes}[para,flushleft]
\scriptsize \textit{*FUTR3D paper provides only general mAP for C+R=0.35. Based on the comparison of  C+L (single-beam) general and car performance, we estimate C+R car class mAP to be somewhere between 0.52-0.54.}
\end{tablenotes}
\end{threeparttable}
\end{table}

The most related similar 3D camera and radar fusion solutions are CenterFusion and FUTR3D. With the final mAP score calculated for car class objects, we see improvements over those two methods in our CDSM model. Additionally, even though our camera-only model achieves a lower score than a similar FCOS3D model, the application of fusion bridges the gap and surpasses both vision-only methods. 

\section{Conclusion}

In this article, we focused on sensor data fusion from camera and radar devices in autonomous vehicle applications. We presented related work for single sensor methods, as well as fusion solutions available for the given sensors suite. On top of that, we thoroughly described our novel approach to the problem with the use of proposed Cross-Domain Spatial Matching transformation and fusion. 

In order to justify CDSM fusion benefits, we conducted experiments on the open NuScenes dataset. We trained both single-sensor models and proposed fusion architecture. The presented results show significant improvements in the latter one, both in general mAP metric and specific corner cases. Finally, we compared our approach to other state-of-the-art solutions in the 3D object detection domain, achieving outstanding performance for the camera and radar setup.

Although we are satisfied with the current results, we observe a gap between the camera and the radar single sensor contribution to the fusion. We believe that applying a machine learning approach to raw radar antennas signal, rather than internally post-processed detections, could improve the perception of these sensors and, with our approach, of the entire fusion system.
\section*{Acknowledgments}
The research was carried out in cooperation
with Aptiv Services Poland S.A. – Technical Center Kraków and AGH University of Krakow.
It is funded partially  by the Polish Ministry of Science and Higher Education (MNiSW) Project No. 0014/DW/2018/02 and partially by National Science Centre, contract no. UMO-2021/41/B/ST7/03851.

\bibliographystyle{elsarticle-num} 
\bibliography{bibliography}
\end{document}